\documentclass[11pt]{article}
\pdfoutput=1

\usepackage[final]{acl}

\usepackage{times}
\usepackage{latexsym}

\usepackage[T1]{fontenc}

\usepackage[utf8]{inputenc}

\usepackage{microtype}

\usepackage{inconsolata}

\usepackage{graphicx}
\usepackage{placeins}
\usepackage{booktabs}
\usepackage{colortbl}
\definecolor{ourscol}{gray}{0.93}
\usepackage{amsmath}
\usepackage{amssymb}
\usepackage{tikz}
\usetikzlibrary{arrows.meta}
\usepackage{algorithm}
\usepackage{algpseudocode}
\title{Persistent Teacher Anchoring for Tool-Using Agents}

\author{
  \textbf{Hyun Bin Park\textsuperscript{1}},
  \textbf{Kyungho Song\textsuperscript{2}},
  \textbf{Sangmin Lee\textsuperscript{1}},
  \textbf{Du-Seong Chang\textsuperscript{1}}
\\
  \textsuperscript{1}Sogang University,
  \textsuperscript{2}University of Michigan, Ann Arbor
\\
  \small{
    \texttt{tjrtkroal@sogang.ac.kr},
    \texttt{songkh@umich.edu}
  }
\\
  \small{
    \texttt{hectorkorea@gmail.com},
    \texttt{duseong.chang@gmail.com}
  }
}

\begin{document}
\maketitle
\begin{abstract}
Distillation is common in LLM post-training, where on-policy knowledge distillation (OPKD) uses student-generated trajectories to prepare the student for downstream RL. At each state, the student matches a next-token distribution supplied by the teacher. As the rollout enters states the teacher would not visit, the teacher-student distribution gap can accumulate. In tool use, this gap becomes consequential because student-written calls execute before supervision and their observations shape later prefixes. Proposer-verifier generation addresses this drift by letting the teacher decide which student-proposed text is retained during generation. Existing formulations govern text but leave tool execution outside their scope. We propose Persistent Teacher Anchoring (PTA), a student-induced but teacher-committed rollout construction. PTA retains chunk-level verification and adds turn-level commitment, allowing a call to reach the environment only after the teacher has verified the entire turn. Treating verified chunks as atomic generation units, we introduce persistent lookahead, which fills idle rollout capacity by advancing future samples and carrying unfinished ones across student updates under the fixed verifier. Across Search-R1-style retrieval and DeepEyes-style perception RL, applying PTA before downstream RL improves macro best@4 by 2.5 and 2.8 points over OPKD under the same downstream RL budget, while lookahead improves throughput by 24\%.
\end{abstract}

\begin{table*}[t]
\centering
\small
\setlength{\tabcolsep}{3.4pt}
\begin{tabular}{@{}lllll@{}}
\toprule
Method & Proposal unit & Rejection trigger & Stage & Decision reaches \\
\midrule
Speculative decoding~\citep{leviathan2023fast} & chunk & probability ratio (lossless) & decoding & text \\
BiLD~\citep{kim2023bild} & token & confidence or disagreement & decoding & text \\
RSD~\citep{liao2025rsd} & reasoning step & step reward below threshold & decoding & text \\
SWITCH~\citep{koo2025switch} & token & teacher-student divergence & training & text \\
SKD~\citep{xu2025speculative} & chunk & outside teacher top-$K$ & training & text \\
SCoRe~\citep{lyu2026score} & full trajectory & teacher-identified error & training & text \\
\rowcolor{ourscol} \textbf{PTA (ours)} & chunk & outside teacher top-$K$ & training & \textbf{text + tool execution} \\
\bottomrule
\end{tabular}
\caption{Proposer-verifier generation methods. A weaker model proposes and a stronger model decides what is kept. The text-only formulations define no execution event, which is what PTA supplies (Section~\ref{sec:rollout-semantics}).}\label{tab:proposer-verifier}
\end{table*}

\section{Introduction}

Large language models increasingly act as tool-using agents, interleaving multi-step reasoning with calls to external tools. Each call acts on the environment and returns an observation, whether retrieved evidence in search or a zoomed image region in visual perception. That observation joins the prefix, and the model generates everything that follows on top of it. Distillation, and on-policy distillation in particular, is often used to shape the initial policy before downstream RL \citep{agarwal2024onpolicy,bai2025qwen3vl}.

On-policy distillation shapes that initial policy on trajectories the student generates itself, so the teacher-student distribution gap can accumulate over a long rollout. In a tool runtime the cost is higher: a deviating call has already executed by the time the teacher signal arrives. Proposer-verifier methods address this accumulation by moving the teacher's decision into generation itself. The weaker model proposes each span, and the stronger model decides what is kept in a single stream of text. These methods are defined for text generation, without a rule for when a generated call may execute. A tool-use rollout does not stay inside that stream. A tool call reads as ordinary text, but its effect lands in the environment, and the observation that returns was produced by neither model. When a proposed call may act on the environment, and how the returned observations are to be treated, are left unspecified across this proposer-verifier line (Section~\ref{sec:related}).

We propose \textbf{Persistent Teacher Anchoring (PTA)}, a pre-RL distillation stage that extends the reach of this decision to the tool calls themselves. The student still proposes every chunk, but a chunk stays only after teacher verification. A turn enters the trajectory and triggers tool execution only once it is finalized as a committed turn. Nothing unverified acts on the environment. Distillation is applied at the committed assistant positions, with observations kept as context rather than imitation targets. Together, these choices give downstream RL an initial policy trained on prefixes that no unverified tool call has shaped.

Long-horizon tool-use rollouts create a structural utilization problem in synchronous training. Samples finish at different times, leaving released slots idle until the slowest rollout completes. We observe that the teacher verifier remains fixed even as the student changes at every update. This suggests that rollout work can continue across updates while remaining teacher-anchored. We therefore make the atomic states produced by teacher anchoring persistent across student updates. Each rollout pauses only after a verified chunk or after the environment has executed a committed tool call and returned its observation. The carried state remains teacher-verified across the student update, and every new chunk from the updated student passes through the same verifier. With that verifier governing both sides of the update, persistent lookahead fills idle slots with future samples and carries unfinished ones into the next update. It changes the schedule, while PTA's verification, commitment, and tool execution remain unchanged.

Our contributions are as follows:
\vspace{-0.4em}
\begin{itemize}
    \setlength{\itemsep}{0pt}
    \setlength{\parskip}{0pt}
    \setlength{\parsep}{0pt}
    \item We propose \textbf{teacher anchoring}, a student-induced but teacher-committed rollout construction: the teacher decides both what enters the trajectory and which tool calls reach the environment.
    \item We make \textbf{teacher anchoring persistent} across student updates: its atomic state boundaries and fixed verifier allow lookahead to fill idle slots and carry unfinished rollout work into the next update, improving scheduler throughput by 24\%.
    \item We validate PTA on \textbf{tool-using agents}: under the same downstream RL budget, it improves macro best@4 over OPKD by 2.5 points in retrieval-mediated reasoning and 2.8 points in perception-mediated reasoning. The advantage is already present at the pre-RL checkpoint and does not survive an ablation that suppresses teacher corrections inside tool-call spans.
\end{itemize}
\vspace{-0.4em}

\input{pta-figure}

\section{Related Work}\label{sec:related}

\subsection{On-Policy Knowledge Distillation}

In on-policy distillation the teacher enters only after the fact. The student generates the trajectory, teacher feedback is computed over it \citep{agarwal2024onpolicy}, and this setup is now standard in post-training pipelines \citep{qwen2025qwen3, glm2026glm5}. This on-policy form descends from knowledge distillation \citep{hinton2015distilling} and avoids the mismatch of off-policy variants, which train the student on teacher-generated sequences it may never produce itself. The teacher's contribution is a distribution. It assigns token-level probabilities over the student's completed sequence, which the student is trained to match, typically under a KL objective. No teacher decision enters the generation itself.

\subsection{Proposer-Verifier Generation}

Proposer-verifier generation places the stronger model's decision inside generation. The pattern began at the decoding stage, where the interaction ends when the verified text is emitted. In speculative decoding, a small model drafts a chunk of tokens and the large model verifies them, fixing which tokens are emitted while its own distribution is exactly preserved \citep{leviathan2023fast}. BiLD relaxes this exactness into a judgment of whether the draft is acceptable, with the large model stepping in when the small model is unconfident or the two disagree \citep{kim2023bild}. RSD widens the unit to a whole reasoning step, where a process reward model scores each proposed step and the strong model takes over a step whose score falls short \citep{liao2025rsd}.

The pattern later entered the training stage, where the verified or corrected sequence is used to update the student. SWITCH lets the teacher intervene at positions where the two models' token probabilities diverge \citep{koo2025switch}. SKD verifies each student-proposed chunk and resamples what it rejects, distilling the student on the resulting sequences \citep{xu2025speculative}. SCoRe brings the pattern to agent trajectories, where the teacher corrects the earliest error in a completed student rollout and the student trains on the corrected trajectory \citep{lyu2026score}. Stage and proposal unit are separate choices in this family. Speculative decoding and SKD apply chunk-level proposal and verification at the decoding and training stages, respectively.

Table~\ref{tab:proposer-verifier} shows that prior decisions reach only text. PTA extends that reach to tool execution. Chunk verification governs what enters the pending assistant turn, and turn commitment governs which tool calls reach the environment.

\noindent\begin{minipage}{\columnwidth}
These boundaries also take on a second role. A verified chunk or a completed environment transition marks a point at which the training rollout may pause and resume. Because the verifier remains fixed, states admitted at these boundaries retain their standing across student updates and form the basis of persistent lookahead (Section~\ref{sec:lookahead-scheduling}).
\end{minipage}

\subsection{Tool-Use Distillation from Constructed Trajectories}

Another line trains the student on constructed trajectories that the student itself did not generate. ToolACE automatically generates and verifies function-calling data \citep{liu2025toolace}. Magnet synthesizes multi-turn trajectories via graph translation \citep{yin2025magnet}. \citet{ibrahim2025finetuning} extract hints from agent failures, generate improved trajectories with retrieval-augmented teachers, and fine-tune students on those trajectories with the hints removed. Supervised fine-tuning on teacher-generated tool-use trajectories is likewise a common initialization before agentic RL. These approaches fix the trajectory before training and take its tokens as the hard target. PTA, in contrast, remains on-policy. The student proposes every assistant span of the rollout it trains on, and the teacher accepts or corrects the proposals rather than writing the trajectory. What the student matches is the teacher's token-level distribution at committed positions, not a sequence of tokens to reproduce.

\subsection{Tool-Use RL}

Tool-use RL trains the agent on task reward alone. Search-R1 learns to issue search queries and use the retrieved evidence \citep{jin2025searchr1}, and DeepEyes induces active perception in multimodal contexts \citep{zheng2026deepeyes}, both by direct RL on the base model. Reward optimization explores on top of the initial policy and inherits the tool-use behavior already encoded in it. PTA is built for this stage rather than for distillation alone. It supplies the initial policy, and we judge it by what a fixed downstream RL budget reaches from that start, in these two environments (Section~\ref{sec:experiments}).

\section{Method}\label{sec:method}

This section defines \textbf{Persistent Teacher Anchoring (PTA)}
as a student-induced but teacher-committed rollout construction
for tool-using agents. We first derive what chunk-based
proposer-verifier generation needs when its text can trigger a
tool. We then define the rollout state, commitment rule, and
distillation objective. The final subsection shows how the
resulting atomic states persist across student updates and support
lookahead scheduling.

\subsection{From Proposer-Verifier Generation to Tool-Use Rollouts}\label{sec:requirements}

Table~\ref{tab:proposer-verifier} separates proposer-verifier
methods by stage. The family began in decoding, where the
interaction ends when the verified text is emitted. It later
entered training, where the verified or corrected sequence is
used to update the student.

PTA builds on the chunk-level proposal and verification used at
both stages. A chunk enters the prefix only after verification.
At the first rejection, the verifier replaces the rejected token,
discards the remaining suffix generated from the old prefix, and
resumes generation from the corrected prefix. The verified chunk
becomes the atomic generation unit used throughout PTA. In a
tool-use rollout, however, the generated sequence can trigger a
tool call and receive an observation from the environment.

A tool-use rollout adds two elements outside the verified text
stream.
\begin{enumerate}
    \setlength{\itemsep}{0pt}
    \setlength{\parskip}{0pt}
    \item \textbf{Calls that act on the environment.}
    A tool call is written as ordinary assistant text, but its
    effect leaves the text stream. Once the call is complete,
    the tool runtime may execute it and the environment returns
    an observation. Applying the text-stream
    procedure directly to such a call creates two distinct
    failure modes.
    (i)~\emph{Execution before verification.} If the runtime
    executes a call as soon as it is syntactically complete, the
    call can reach the environment before the teacher has
    finished verifying the turn that contains it
    (Figure~\ref{fig:pta-rollout-overview} (a)).
    (ii)~\emph{Correction after execution.} The returned
    observation is conditioned on the call exactly as it was
    executed. Once the call has executed, revising any token in
    that call or in the prefix that precedes it would leave the
    revised sequence inconsistent with the returned observation.
    Verification must therefore finish before the call is allowed
    to execute.
    \item \textbf{Observations written back by the environment.}
    After execution, the environment returns an observation,
    which the tool runtime inserts into the same sequence used
    for subsequent generation. Because neither model produced
    these tokens, their presence in the verified sequence creates
    two further gaps.
    (iii)~\emph{Ungoverned spans.} The inherited verification
    rule can keep a student proposal or replace it with teacher
    content. An observation fits neither branch.
    (iv)~\emph{Undefined targets.} A distillation objective over
    the resulting sequence also needs a rule for whether
    observation tokens are imitation targets. Treating every
    token alike would train the student to reproduce content
    supplied by the environment.
\end{enumerate}

Calls therefore need a second decision beyond what text is kept.
PTA retains chunk-level verification and adds turn-level
commitment. Each verified chunk is appended to a pending assistant
turn, which remains outside the committed trajectory. When the
turn reaches an assistant EOS or a tool-call closing tag, PTA
finalizes it and appends the whole turn to the trajectory as a
committed assistant turn. Only after commitment does the tool
runtime parse the turn and, if it is a valid tool call, execute it.
Turn-level commitment thus decides when verified text may act on
the environment (Figure~\ref{fig:pta-rollout-overview} (b)).

Observations need a source rule instead. PTA distinguishes prompt
spans, committed assistant turns, and environment observations.
All three remain in the prefix for later generation, while the
distillation loss applies only at committed assistant positions.
After a committed tool call executes, its observation is appended
before generation resumes. Sections~\ref{sec:rollout-semantics}
and~\ref{sec:objective} formalize this construction and its
objective.

\subsection{Teacher-Anchored Rollout Construction}\label{sec:rollout-semantics}

At interaction turn $t$, the rollout state is
$S_{t,k}=(\tau_t,R_{t,k})$. The \textbf{committed trajectory}
$\tau_t$ contains prompt spans, committed assistant turns, and
environment observations. The \textbf{pending assistant turn}
$R_{t,k}$ contains the $k$ verified chunks accumulated for the
next assistant turn, with $R_{t,0}=\emptyset$. Both components
condition the next proposal, but $R_{t,k}$ remains outside
$\tau_t$. Only a turn appended to $\tau_t$ may trigger an
environment transition.

Given $\tau_t$ and $R_{t,k-1}$, the student proposes a chunk
$\tilde{c}_{t,k}$. Let $V_\phi$ denote the teacher verifier. It
accepts or corrects the proposal and returns
$c^{\mathrm{v}}_{t,k}=V_\phi(\tilde{c}_{t,k}\mid\tau_t,R_{t,k-1})$.
Verification proceeds token by token under the verified prefix,
which consists of $\tau_t$, $R_{t,k-1}$, and the tokens already
verified in the current chunk. A student token is kept when it
lies in the teacher's top-$K$ support. At the first rejection,
the verifier replaces that token with the teacher's top-1 token
and discards the remaining suffix because it was generated from
a prefix that has now changed. The student then proposes a new
chunk from the corrected prefix.

\begin{samepage}
PTA appends the verified chunk in one state transition,
\[
(\tau_t,R_{t,k-1})
\xrightarrow{V_\phi}
(\tau_t,R_{t,k-1}\oplus c^{\mathrm v}_{t,k}).
\]
\end{samepage}
Neither the raw proposal nor its discarded suffix appears in the
next rollout state. The verified chunk is therefore the smallest
unit that PTA appends to a pending turn.

When the pending turn reaches an assistant EOS or a tool-call
closing tag, PTA finalizes it as the committed assistant turn
$A_t$ and appends the whole turn to $\tau_t$. Only after
commitment does the tool runtime parse $A_t$. A final answer
terminates at this point. If $A_t$ is a valid tool call, the
runtime executes it, receives the observation $o_t$, and appends
$o_t$ as a separate observation span before generation resumes.
The next exposed state is therefore
\[
S_t^+=
\begin{cases}
(\tau_t\oplus A_t,\emptyset),
& \text{final answer}, \\
(\tau_t\oplus A_t\oplus o_t,\emptyset),
& \text{valid tool call}.
\end{cases}
\]
PTA exposes a rollout state only after chunk verification or after
a complete environment transition, never at an intermediate point.
In particular, no resumable state contains a raw speculative chunk,
a partial assistant turn in the committed trajectory, or a committed
tool call awaiting its observation. The outer rollout ends after a
final answer or when its budget is exhausted.

\begin{algorithm}[b]
  \footnotesize
  \caption{PTA rollout with chunk-level verification and turn-level commitment}\label{alg:pta-rollout}
  \begin{algorithmic}[1]
    \State{} Initialize committed trajectory $\tau$
    \While{$\tau$ is not terminal \textbf{and} budget remains}
    \State{} $R \gets \emptyset$ \Comment{pending assistant turn}
    \While{$R$ is not finalized}
    \State{} $\tilde{c} \gets \mathrm{Student.propose\_chunk}(\tau, R)$
    \State{} $c^{\mathrm{v}} \gets \mathrm{Teacher.verify\_chunk}(\tau, R, \tilde{c})$
    \State{} $R \gets R \oplus c^{\mathrm{v}}$ \Comment{verified boundary}
    \EndWhile{}
    \State{} $A \gets \mathrm{finalize}(R)$ \Comment{committed assistant turn}
    \State{} $\tau \gets \tau \oplus A$
    \State{} $u \gets \mathrm{parse\_tool\_call}(A)$
    \If{$u$ is valid}
    \State{} $o \gets \mathrm{Environment.step}(u)$
    \State{} $\tau \gets \tau \oplus o$ \Comment{transition complete}
    \ElsIf{$A$ is a final answer}
    \State{} mark $\tau$ as terminal
    \EndIf{}
    \EndWhile{}
  \end{algorithmic}
\end{algorithm}

The same construction fits high-throughput rollout engines that
generate assistant continuations in chunks, advance many samples
in parallel, and reuse partial prefixes. We implement PTA in
\texttt{verl} with \texttt{sglang} as the inference engine and
insert teacher verification into its chunk-level generation loop.
Section~\ref{sec:lookahead-scheduling} reuses these state
boundaries without changing verification, commitment, or tool
execution.

\subsection{Source-Aware Objective}\label{sec:objective}

PTA applies distillation only to committed assistant positions.
Let the committed trajectory for sample $b$ be
$\tau_b=(x_1,\ldots,x_n)$, where each token inherits a source
label marking it as prompt, assistant, or observation content.
The target positions are
\[
\mathcal{A}(\tau_b)
=
\{i \mid \mathrm{src}(x_i)=\mbox{\textsc{Assistant}}\}.
\]
Because a pending turn remains outside $\tau_b$, its tokens enter
this set only after commitment. Prompt and observation tokens
remain in the prefix but are not imitation targets. Treating
observations as targets would train the student to reproduce
content supplied by the environment.

For training batch $\mathcal{B}_t$, let $q_{b,i}$ and $p_{b,i}$
denote the teacher and student next-token distributions at
committed assistant position $i$ of $\tau_b$. PTA minimizes
\[
\mathcal{L}_{\mathrm{PTA}}(\theta)
=
\frac{1}{N_t}
\sum_{b\in\mathcal{B}_t}
\sum_{i\in\mathcal{A}(\tau_b)}
D_{\mathrm{KL}}(q_{b,i}\|p_{b,i}),
\]
where $N_t=\sum_{b\in\mathcal{B}_t}|\mathcal{A}(\tau_b)|$.
PTA keeps the token-level distribution-matching loss unchanged.
It changes the rollout and the positions to which the loss is
applied.

%
%
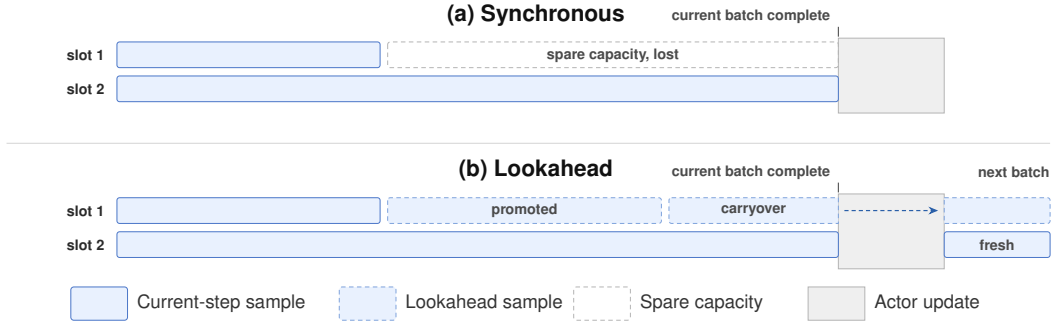
\begin{figure*}[t]
\centering
\begin{tikzpicture}[x=1mm,y=-1mm]
\useasboundingbox (5,0) rectangle (139,43);

\node[anchor=center,inner sep=0pt,text=pta111111] at (70,2.6) {\ptaTtl{(a) Synchronous}};
\node[anchor=east,inner sep=0pt,text=pta4A4A4A] at (109,2.8) {\ptaLbl{current batch complete}};
\path[fill=ptaEFEFEF,draw=ptaB4B4B4,line width=0.26pt] (110,5.7) rectangle ++(14,9.9);
\node[anchor=east,inner sep=0pt,text=pta2A2A2A] at (13,7.9) {\ptaLbl{slot 1}};
\node[anchor=east,inner sep=0pt,text=pta2A2A2A] at (13,12.4) {\ptaLbl{slot 2}};
\path[fill=ptaE9F1FE,draw=pta4A76C8,line width=0.28pt,rounded corners=0.35mm] (14.6,6.2) rectangle ++(34.8,3.4);
\path[fill=ptaE9F1FE,draw=pta4A76C8,line width=0.28pt,rounded corners=0.35mm] (14.6,10.7) rectangle ++(95.4,3.4);
\path[draw=ptaB4B4B4,line width=0.26pt,dash pattern=on 0.6mm off 0.45mm,rounded corners=0.35mm] (50.4,6.2) rectangle ++(59.6,3.4);
\node[anchor=center,inner sep=0pt,text=pta4A4A4A] at (80.2,7.9) {\ptaLbl{spare capacity, lost}};
\draw[draw=pta4A4A4A,line width=0.4pt] (110,4.2) -- (110,5.5);

\path[draw=ptaD4D4D4,line width=0.3pt] (0,19.6) rectangle ++(138,0);

\node[anchor=center,inner sep=0pt,text=pta111111] at (70,23.2) {\ptaTtl{(b) Lookahead}};
\node[anchor=east,inner sep=0pt,text=pta4A4A4A] at (109,23.4) {\ptaLbl{current batch complete}};
\node[anchor=east,inner sep=0pt,text=pta4A4A4A] at (138,23.4) {\ptaLbl{next batch}};
\path[fill=ptaEFEFEF,draw=ptaB4B4B4,line width=0.26pt] (110,26.3) rectangle ++(14,9.9);
\node[anchor=east,inner sep=0pt,text=pta2A2A2A] at (13,28.5) {\ptaLbl{slot 1}};
\node[anchor=east,inner sep=0pt,text=pta2A2A2A] at (13,33.0) {\ptaLbl{slot 2}};
\path[fill=ptaE9F1FE,draw=pta4A76C8,line width=0.28pt,rounded corners=0.35mm] (14.6,26.8) rectangle ++(34.8,3.4);
\path[fill=ptaE9F1FE,draw=pta4A76C8,line width=0.28pt,rounded corners=0.35mm] (14.6,31.3) rectangle ++(95.4,3.4);
\path[fill=ptaE9F1FE,draw=pta7FA3DC,line width=0.32pt,dash pattern=on 0.6mm off 0.45mm,rounded corners=0.35mm] (50.4,26.8) rectangle ++(36.2,3.4);
\node[anchor=center,inner sep=0pt,text=pta4A4A4A] at (68.5,28.5) {\ptaLbl{promoted}};
\path[fill=ptaE9F1FE,draw=pta7FA3DC,line width=0.32pt,dash pattern=on 0.6mm off 0.45mm,sharp corners]
  (87.6,26.8) -- (110,26.8) -- (110,30.2) [rounded corners=0.35mm] -- (87.6,30.2) -- cycle;
\node[anchor=center,inner sep=0pt,text=pta4A4A4A] at (98.8,28.5) {\ptaLbl{carryover}};
\path[fill=ptaE9F1FE,draw=pta7FA3DC,line width=0.32pt,dash pattern=on 0.6mm off 0.45mm,rounded corners=0.35mm]
  (124,26.8) -- (138,26.8) -- (138,30.2) [sharp corners] -- (124,30.2) -- cycle;
\draw[-{Stealth[length=1.3mm,width=1.05mm]},draw=pta2A5FA8,line width=0.45pt,dash pattern=on 0.55mm off 0.4mm] (110.8,28.5) -- (123.2,28.5);
\path[fill=ptaE9F1FE,draw=pta4A76C8,line width=0.28pt,rounded corners=0.35mm] (124,31.3) rectangle ++(14,3.4);
\node[anchor=center,inner sep=0pt,text=pta4A4A4A] at (131,33.0) {\ptaLbl{fresh}};
\draw[draw=pta4A4A4A,line width=0.4pt] (110,24.8) -- (110,26.1);

\path[fill=ptaE9F1FE,draw=pta4A76C8,line width=0.3pt,rounded corners=0.4mm] (8.5,38.6) rectangle ++(7.5,4.4);
\node[anchor=west,inner sep=0pt,text=pta333333] at (17.2,40.8) {\ptaLeg{Current-step sample}};
\path[fill=ptaE9F1FE,draw=pta7FA3DC,line width=0.3pt,dash pattern=on 0.6mm off 0.45mm,rounded corners=0.4mm] (44,38.6) rectangle ++(7.5,4.4);
\node[anchor=west,inner sep=0pt,text=pta333333] at (52.7,40.8) {\ptaLeg{Lookahead sample}};
\path[draw=ptaB4B4B4,line width=0.3pt,dash pattern=on 0.6mm off 0.45mm,rounded corners=0.4mm] (75,38.6) rectangle ++(7.5,4.4);
\node[anchor=west,inner sep=0pt,text=pta333333] at (83.7,40.8) {\ptaLeg{Spare capacity}};
\path[fill=ptaEFEFEF,draw=ptaB4B4B4,line width=0.3pt] (106,38.6) rectangle ++(7.5,4.4);
\node[anchor=west,inner sep=0pt,text=pta333333] at (114.7,40.8) {\ptaLeg{Actor update}};
\end{tikzpicture}
\caption{Schematic of synchronous and lookahead scheduling over the same
wall-clock interval. Ticks mark completion of the current-step batch. In (b),
one lookahead sample is promoted before the boundary, another becomes carryover
and resumes after the actor update, and a fresh sample fills the remaining slot
in the next batch. The carryover cut occurs only at an atomic state boundary.}
\label{fig:lookahead-schedule}
\end{figure*}

\subsection{Persistent Lookahead Scheduling}\label{sec:lookahead-scheduling}

Long-horizon tool-use rollouts finish at different times. Some
samples reach a final answer quickly, while others wait through
multiple tool calls, long observations, or repeated teacher
verification. A synchronous PTA step waits for every sample in
its current-step batch, so slots released by earlier completions
remain idle until the slowest sample finishes. PTA can turn this
lost capacity into future rollout work because the student and
teacher have different lifetimes. The student changes at every
update, while the teacher verifier remains fixed throughout
training.

Section~\ref{sec:rollout-semantics} exposes a rollout state only
after a verified chunk or a complete environment transition. These
atomic boundaries determine where a rollout may pause. Because the
verifier is fixed, a state it has admitted keeps its standing after
a student update. The update changes the student, not the carried
state, and every new chunk must pass the same verifier. Any tool
call already present in the state was committed before execution
and is accompanied by its returned observation.

\textbf{Lookahead scheduling} fills each released slot with a
future sample while the current-step batch is still running. The
current-step batch defines the lookahead frontier. A lookahead
sample that finishes before the current-step batch is complete is
\textbf{promoted} into the current update batch, and its slot can
accept another lookahead sample. When every current-step sample has
finished, the scheduler stops launching new lookahead work.
Lookahead samples still in progress become \textbf{carryover} for
the next step. Figure~\ref{fig:lookahead-schedule} shows this
lifecycle. The step does not become shorter. Lookahead fills time
that the synchronous step leaves unused. It changes when a sample
is advanced, not how it is verified, committed, or executed.

The cutoff never interrupts an atomic unit already in progress.
A sample with an unfinished assistant turn pauses only after its
current chunk has been verified and appended. If the verified
chunk completes a tool-call turn, PTA commits the turn, executes
the call, and appends the observation before saving the state.
Every carryover sample is therefore a lookahead sample paused at a
state boundary from Section~\ref{sec:rollout-semantics}.
Current-step samples cannot become carryover because their
completion defines the cutoff.

The update batch contains the completed current-step batch and
all promoted samples. The scheduler forms the next current-step
batch from the carryover samples and enough fresh dataset samples
to restore its nominal size. The updated student resumes each
carryover state, and every newly generated chunk again passes the
fixed teacher verifier. Appendix~\ref{app:teacher-gated-carryover}
formalizes why this continuation preserves PTA commitment semantics
across student updates.

\section{Experiments}\label{sec:experiments}

We evaluate PTA across retrieval and perception tools under a matched downstream RL budget. We then ask whether the difference is already present before RL and how the same rollout construction affects sparse distillation and scheduler throughput.

\begin{table*}[t]
\centering
\begin{minipage}[t]{0.49\textwidth}
\centering
\scriptsize
\setlength{\tabcolsep}{3pt}
\scalebox{1.14}{%
\begin{tabular}{lrrr>{\columncolor{ourscol}}r}
\toprule
Metric & Base & Direct RL & OPKD + RL & PTA + RL \\
\midrule
\multicolumn{5}{c}{\textbf{Benchmark-level results}} \\
\midrule
NQ mean@4 & 24.04 & \textbf{29.63} & 29.04 & 29.45 \\
NQ best@4 & 36.84 & 38.34 & 38.76 & \textbf{39.09} \\
PopQA mean@4 & 31.74 & \textbf{36.45} & 33.88 & 35.49 \\
PopQA best@4 & 45.61 & \textbf{47.62} & 42.83 & 47.12 \\
HotpotQA mean@4 & 17.66 & 21.51 & \textbf{22.13} & 21.98 \\
HotpotQA best@4 & 31.25 & 32.30 & 32.77 & \textbf{33.37} \\
Musique mean@4 & 4.39 & 6.44 & \textbf{8.20} & 10.93 \\
Musique best@4 & \textbf{9.37} & 12.40 & 13.94 & 18.78 \\
\midrule
\multicolumn{5}{c}{\textbf{Aggregate results}} \\
\midrule
Weighted mean@4 & 24.58 & \textbf{28.95} & 27.87 & \textbf{28.95} \\
Weighted best@4 & 37.46 & 39.24 & 37.09 & \textbf{39.92} \\
Macro mean@4 & 19.46 & 23.51 & 23.31 & \textbf{24.46} \\
Macro best@4 & 30.77 & 32.66 & 32.07 & \textbf{34.59} \\
\bottomrule
\end{tabular}}
\caption{Retrieval-mediated reasoning results.}\label{tab:retrieval-results}
\end{minipage}
\hfill
\begin{minipage}[t]{0.49\textwidth}
\centering
\scriptsize
\setlength{\tabcolsep}{3pt}
\scalebox{1.14}{%
\begin{tabular}{lrrr>{\columncolor{ourscol}}r}
\toprule
Metric & Base & Direct RL & OPKD + RL & PTA + RL \\
\midrule
\multicolumn{5}{c}{\textbf{Benchmark-level results}} \\
\midrule
VStar mean@4 & 59.03 & 61.65 & 59.55 & \textbf{64.53} \\
VStar best@4 & 75.39 & 73.30 & \textbf{80.10} & 78.01 \\
H4 mean@4 & 62.88 & 65.25 & 62.25 & \textbf{66.63} \\
H4 best@4 & 64.50 & \textbf{70.00} & 65.50 & 68.50 \\
H8 mean@4 & 54.13 & 58.88 & 53.00 & \textbf{59.63} \\
H8 best@4 & 61.00 & 61.00 & 56.00 & \textbf{63.50} \\
\midrule
\multicolumn{5}{c}{\textbf{Aggregate results}} \\
\midrule
Macro mean@4 & 58.68 & 61.92 & 58.27 & \textbf{63.59} \\
Macro best@4 & 66.96 & 68.10 & 67.20 & \textbf{70.00} \\
Weighted mean@4 & 58.56 & 62.02 & 57.83 & \textbf{63.27} \\
Weighted best@4 & 64.10 & 66.33 & 62.81 & \textbf{67.28} \\
\bottomrule
\end{tabular}}
\caption{Benchmark-level perception-mediated reasoning results.}\label{tab:perception-results}
\end{minipage}
\end{table*}

\subsection{Experimental Setup}

\subsubsection{Environments}

We evaluate PTA in two tool-mediated reasoning environments. Their tool interfaces and result modalities differ, but in both environments model-generated calls determine the observations available to later reasoning.

\paragraph{Retrieval-mediated reasoning.}
The model calls a search tool and reasons over the retrieved text, following Search-R1~\citep{jin2025searchr1}. NQ~\citep{kwiatkowski2019natural} and PopQA~\citep{mallen2023popqa} test short factual retrieval, while HotpotQA~\citep{yang2018hotpotqa} and Musique~\citep{trivedi2022musique} require evidence to be gathered and connected across multiple steps.

\paragraph{Perception-mediated reasoning.}
The model calls a bbox-based perception tool and reasons over the returned visual observation, following DeepEyes~\citep{zheng2026deepeyes}. VStar~\citep{wu2024vstar} tests fine-grained attribute and spatial reasoning. HRBench4K and HRBench8K~\citep{wang2025hrbench} test tool use over high-resolution visual contexts, with the two variants distinguishing the input resolution.

\subsubsection{Compared Methods}

We compare four settings. \textbf{Base} is the starting model with the same tool interface used by the trained systems. \textbf{Direct RL} applies the downstream RL objective without an intermediate distillation stage. This matches the established training pattern in Search-R1 and DeepEyes, which apply RL without a supervised cold start.

The main comparison is between two pre-RL alignment settings. \textbf{OPKD + RL} distills on student-generated, student-committed rollouts before downstream RL. \textbf{PTA + RL} uses the same student to propose every chunk, while the teacher verifier determines which assistant turns are committed, and then applies the same downstream RL procedure. The two settings therefore share student-induced rollout generation and the downstream RL budget. They differ in whether teacher information enters after the student has committed the trajectory or participates in commitment itself. Initializations based on teacher-generated trajectories, including supervised fine-tuning and off-policy distillation, change both the trajectory source and the supervision mechanism and are outside this comparison.

In the retrieval-mediated environment, the teacher is Qwen3\mbox{-}32B and the student is Qwen3\mbox{-}1.7B\@; in the perception-mediated environment, the teacher is Qwen3-VL\mbox{-}32B and the student is Qwen3-VL\mbox{-}2B\@. All four settings use the same teacher-student pair within each environment.

\subsection{Retrieval-Mediated Reasoning}\label{sec:retrieval}

Table~\ref{tab:retrieval-results} reports retrieval-mediated reasoning results. All scores are EM percentages.

PTA + RL gives the strongest retrieval aggregates. It leads on weighted best@4, macro mean@4, and macro best@4, and ties Direct RL on weighted mean@4. Its macro best@4 exceeds OPKD + RL by 2.52 points. Direct RL remains competitive on the short factual benchmarks, while PTA + RL leads on best@4 for three of the four datasets. Short factual questions can often be resolved with a brief retrieval path, whereas errors in multi-step tasks continue to shape later queries and evidence.

The largest gain over OPKD + RL appears on Musique, where mean@4 rises from 8.20 to 10.93 and best@4 from 13.94 to 18.78. Musique makes early searches and retrieved evidence consequential because later steps must connect evidence gathered along the way. The concentration of the gain on this benchmark is consistent with PTA preventing unverified interactions from shaping the subsequent retrieval path.

\subsection{Perception-Mediated Reasoning}\label{sec:perception}

Table~\ref{tab:perception-results} reports benchmark-level perception results. Appendix~\ref{app:perception-details} provides VStar subtask and HRBench split results.

PTA + RL leads on all four aggregate metrics. Relative to OPKD + RL, it gains 5.32 points on macro mean@4 and 2.80 points on macro best@4. The advantage therefore extends from retrieved text to tool-returned visual observations.

The benchmark-level results also show where the advantage is less uniform. VStar separates average sampled performance from the strongest of four rollouts. PTA + RL leads in mean@4, while OPKD + RL is higher in best@4. On HRBench4K, PTA + RL leads in mean@4 but Direct RL leads in best@4. PTA + RL leads in both metrics on HRBench8K. The gain is therefore strongest in the higher-resolution setting, where the perception tool must localize evidence in a denser visual context.

\subsection{Alignment before Downstream RL}\label{sec:alignment-stage}

\begin{table*}[t]
\centering
\small
\begin{tabular}{lcccccc}
\toprule
& \multicolumn{2}{c}{VStar} & \multicolumn{2}{c}{HRBench4K} & \multicolumn{2}{c}{HRBench8K} \\
\cmidrule(lr){2-3}\cmidrule(lr){4-5}\cmidrule(lr){6-7}
Setting & mean@4 & best@4 & mean@4 & best@4 & mean@4 & best@4 \\
\midrule
OPKD & 57.85 & 78.53 & 61.00 & 64.00 & 54.63 & 58.00 \\
\rowcolor{ourscol} PTA & \textbf{61.13} & \textbf{81.15} & \textbf{64.00} & \textbf{68.50} & \textbf{57.88} & \textbf{65.00} \\
PTA without tool-call replacement & 50.00 & 70.68 & 55.50 & 58.00 & 48.13 & 50.50 \\
\bottomrule
\end{tabular}
\caption{Pre-RL alignment checkpoints in the perception-mediated environment.}\label{tab:alignment-checkpoint}
\end{table*}

The downstream results leave open whether the difference is already present before reward optimization. Table~\ref{tab:alignment-checkpoint} evaluates the alignment checkpoints at the same training step, before downstream RL. PTA leads OPKD on all six metrics, with the largest gap on HRBench8K best@4, which rises from 58.00 to 65.00. The advantage is therefore already present in the policy handed to reward optimization.

The third row isolates correction inside tool-call spans. It retains full-turn buffering, delayed execution, observation masking, teacher verification, and KL supervision, along with the same threshold, chunk size, and training step. The only change suppresses a teacher replacement when the rejected token lies inside a parsed tool call, so the environment receives the student's original call. Performance falls below both PTA and OPKD on every metric. Safe timing alone is therefore insufficient when the call that reaches the environment can retain rejected content.

The same runs retain the student's proposal for 90.5\% of committed assistant tokens in retrieval and 88.4\% in perception. Replacements inside tool-call spans account for only 0.05\% and 0.82\% of all committed assistant tokens, yet removing them eliminates the alignment advantage. PTA thus remains predominantly student-proposed, while the rare corrections inside tool calls act on content that reaches the environment. Appendix~\ref{app:verifier-stats} reports the full verifier statistics.

\subsection{\texorpdfstring{Top-$N$ Distribution Stability}{Top-N Distribution Stability}}\label{sec:topn}

To examine the distillation conditions behind these checkpoints, we measure how much teacher probability a fixed top-$N$ support preserves at each committed assistant position. Top-$N$ distillation is a standard sparse-logit approximation that stores only the $N$ highest-probability tokens from the teacher distribution~\citep{shum2024first,anshumann2025sparse}. Its fidelity depends on this coverage, which we call the \textit{teacher retained mass}
\begin{equation}
m_N(h) = \sum_{x \in \mathcal{T}_N(h)} q_\phi(x \mid h),
\end{equation}
where $\mathcal{T}_N(h)$ is the teacher's top-$N$ support at position $h$, measured before renormalization. Values near one indicate that the sparse approximation preserves most of the distribution.

\begin{figure*}[t]
\centering
\begin{minipage}{0.30\textwidth}
\centering
\includegraphics[width=\linewidth]{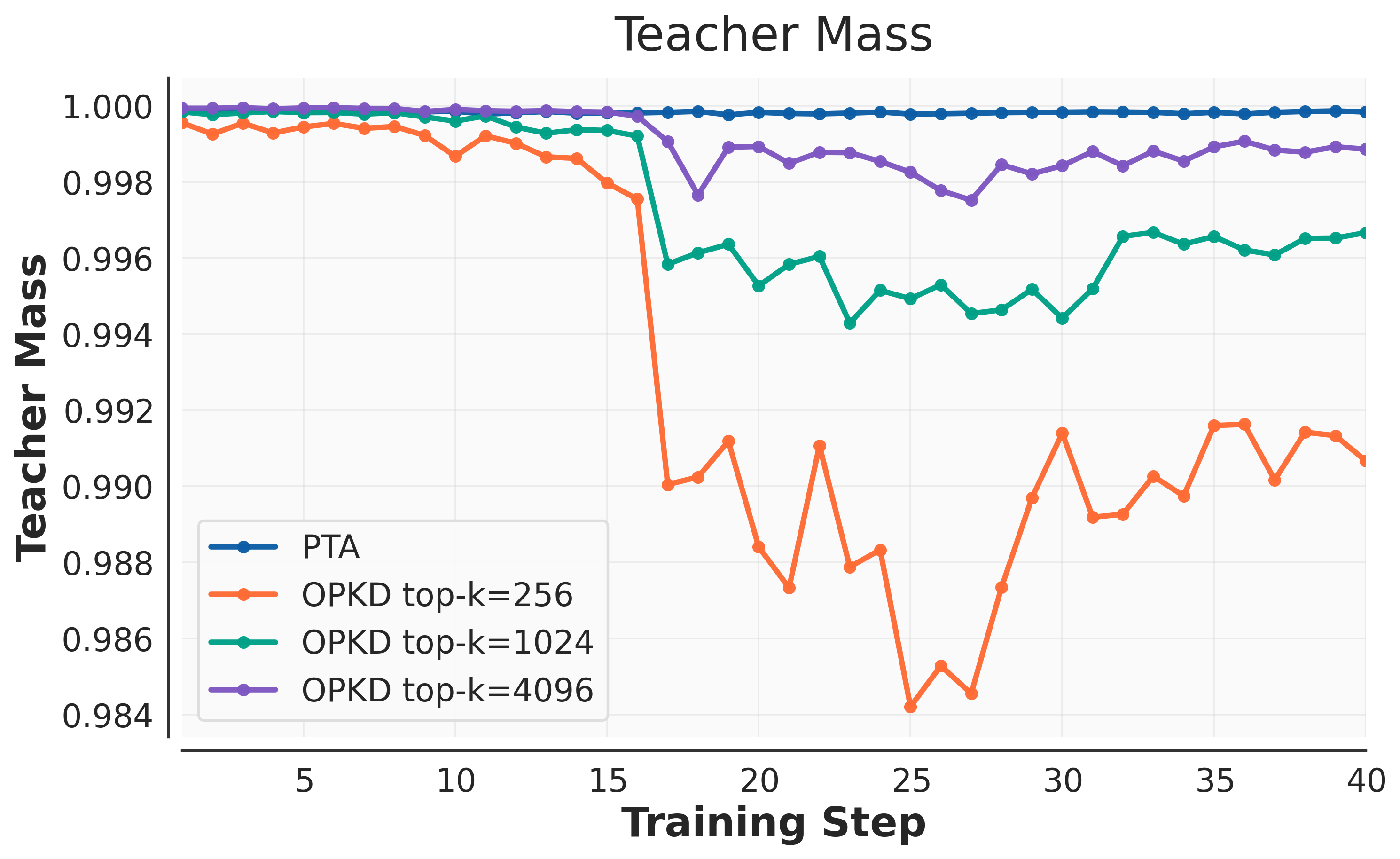}\\[-0.5ex]
{\footnotesize (a) Teacher retained mass}
\end{minipage}
\hfill
\begin{minipage}{0.30\textwidth}
\centering
\includegraphics[width=\linewidth]{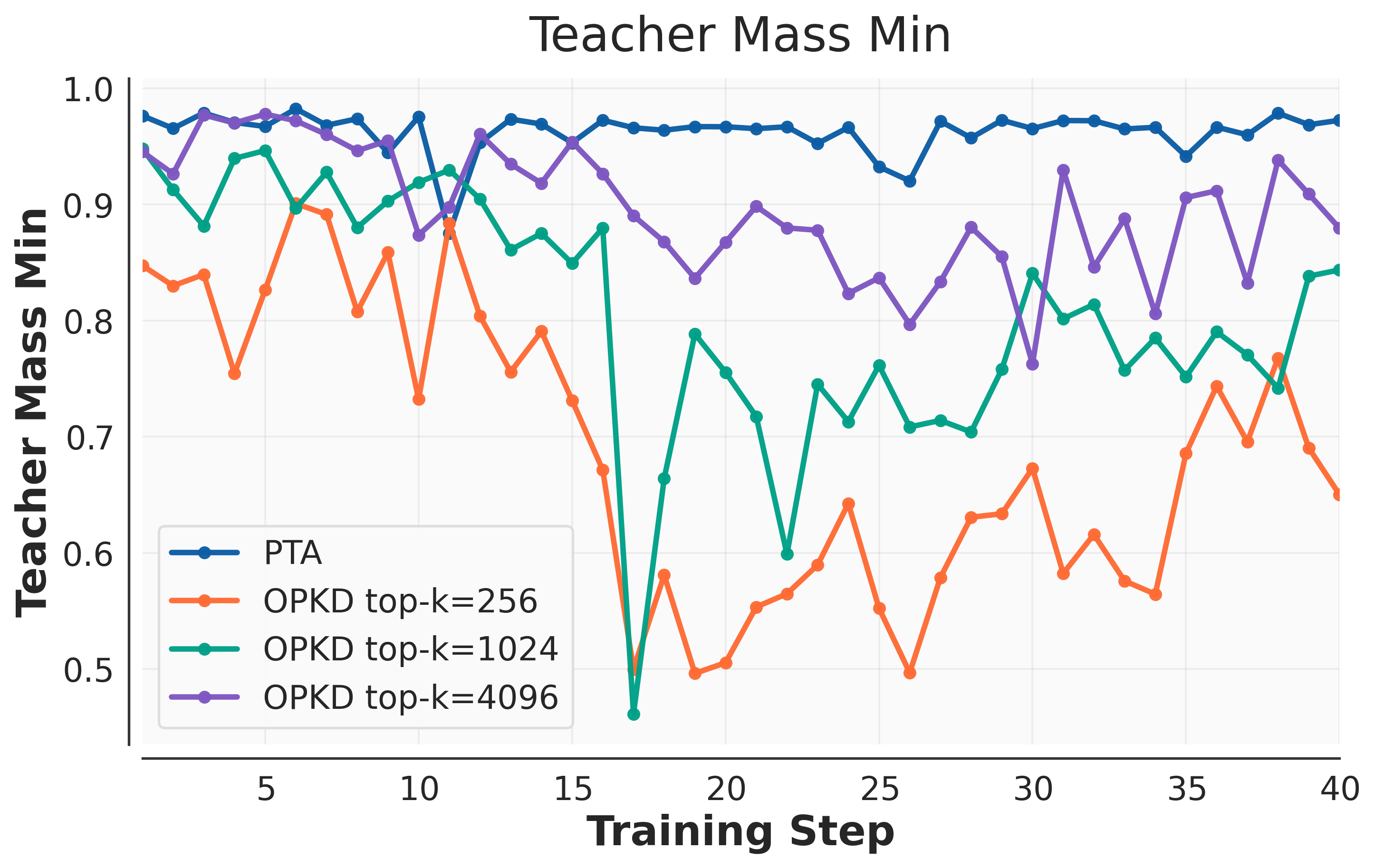}\\[-0.5ex]
{\footnotesize (b) Minimum retained mass}
\end{minipage}
\hfill
\begin{minipage}{0.30\textwidth}
\centering
\includegraphics[width=\linewidth]{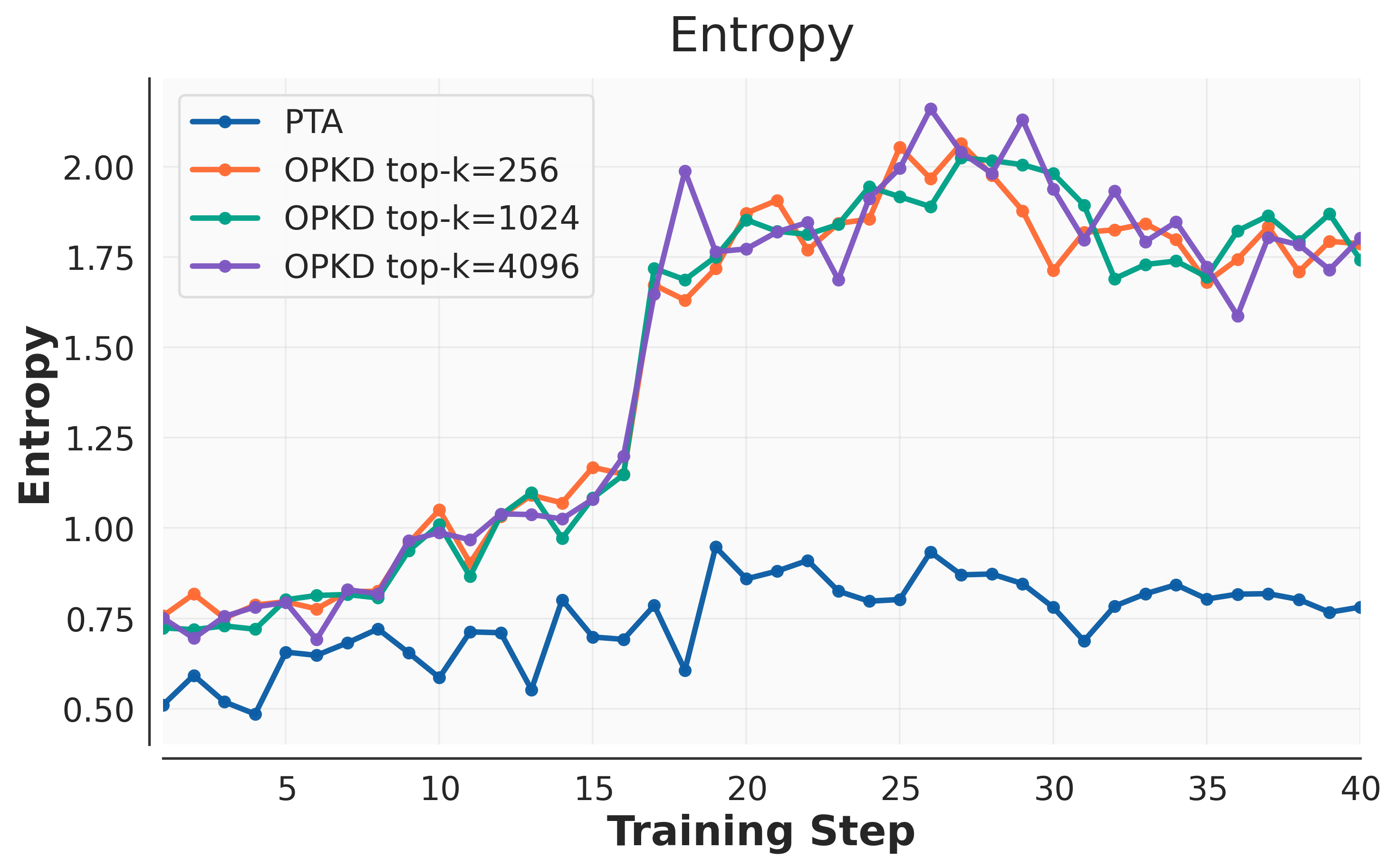}\\[-0.5ex]
{\footnotesize (c) Actor entropy}
\end{minipage}
\caption{Top-N stability diagnostics over an early training window.}\label{fig:topn-stability}
\end{figure*}

Figure~\ref{fig:topn-stability}~(a) shows that PTA maintains high average retained mass with $N=256$, while OPKD needs a larger support to recover comparable coverage. Panel~(b) gives a stricter view through the minimum retained mass. PTA keeps its lowest-coverage positions stable at $N=256$, whereas OPKD exhibits sharp drops at smaller $N$ and still contains low-coverage positions at $N=1024$. Under the same sparse-logit budget, teacher-committed prefixes therefore make the teacher distribution easier to preserve than student-committed prefixes.

Panel~(c) shows that the coverage gap follows a difference in the distribution rather than the truncation rule alone. Actor entropy under OPKD rises as its retained mass falls, while under PTA it remains lower and more stable. Under a fixed top-$N$ budget, OPKD therefore leaves more teacher mass outside the stored support at both typical and worst-case positions.

Together with the checkpoint results, these diagnostics connect rollout construction to the policy passed to downstream RL. On PTA's teacher-committed prefixes, a practical sparse budget preserves more of the teacher distribution before reward optimization begins.

\subsection{Lookahead Efficiency}\label{sec:lookahead-efficiency}

We finally measure the scheduler benefit of persistence with a fixed current-step batch size $B=32$. Throughput is
\begin{equation}
T_{\mathrm{sched}} = \frac{B + P}{t^{\mathrm{gen}} + t^{\mathrm{update}}},
\end{equation}
where $P$ is the number of promoted lookahead samples. The numerator counts the current-step batch and future samples completed in the same wall-clock window. Carryover samples enter the next step and are not counted.

\begin{table}[t]
\centering
\small
\setlength{\tabcolsep}{5pt}
\begin{tabular}{lcc}
\toprule
Setting & Promoted & Throughput \\
\midrule
synchronous & 0.0 & 0.519 \\
lookahead & 20.2 & 0.644 \\
\bottomrule
\end{tabular}
\caption{Lookahead scheduling efficiency.}\label{tab:lookahead-efficiency}
\end{table}

Table~\ref{tab:lookahead-efficiency} shows that lookahead raises throughput from 0.519 to 0.644 samples/s, a 24\% increase, while promoting 20.2 samples per step on average. The scheduler gains throughput by filling otherwise idle rollout capacity, not by shortening generation or update time.

\section{Conclusion}

Persistent Teacher Anchoring extends proposer-verifier generation from deciding what text is retained to deciding when a completed assistant turn may act on the environment. The student proposes each chunk, the teacher verifies it, and only a finalized turn is committed and allowed to execute a tool. These atomic boundaries also make verified rollout states persistent across student updates. Because the teacher verifier remains fixed, lookahead scheduling can pause and resume future samples without changing verification, commitment, or tool execution.

The experiments trace this construction from alignment to downstream RL. At the alignment checkpoint, PTA leads OPKD on all six perception metrics, and the advantage does not survive suppressing corrections inside tool-call spans. With the same downstream RL budget, PTA + RL then improves macro best@4 by 2.5 points on retrieval-mediated reasoning and 2.8 points on perception-mediated reasoning. On teacher-committed prefixes, a fixed top-$N$ budget retains more teacher mass and yields a more faithful sparse approximation. Actor entropy shows the same separation. Persistence also supports lookahead, which improves scheduler throughput by 24\%. PTA thus gives reward optimization an initial policy whose tool-use trajectories have already been shaped by teacher-verified interactions.

\section*{Limitations}

PTA is evaluated in retrieval-mediated and perception-mediated reasoning, which cover textual and visual tool observations. The controlled comparison fixes the student proposer and downstream RL budget, and varies whether teacher information enters before or after commitment. The resulting empirical claim is PTA's advantage over OPKD under these matched conditions. Code execution, database access, and multi-agent interaction are not covered by this study.

PTA intentionally corrects verifier-rejected tool calls before execution. The pre-RL policy therefore does not encounter the error states those calls would have created. Downstream RL restores unrestricted student-driven rollouts. This staged design separates pre-commit correction from later exposure to deployment-like states, while the evaluation targets final policy quality rather than recovery-specific capability. PTA also requires teacher inference during rollout construction. Lookahead uses otherwise idle capacity but does not remove the verifier cost, whose trade-off may change with model scale, tool latency, and task horizon.

The experiments use one specified verifier configuration, with a fixed threshold, chunk size, deterministic top-1 replacement, and verification over all assistant tokens. Teacher-distribution sampling and tool-call-only verification are alternative instantiations rather than factors varied in this study. Each setting is trained once, so the significance analysis characterizes evaluation variation rather than training-seed variation.


\section*{Acknowledgments}

This research was supported by the Ministry of Science and ICT (MSIT), Republic of Korea, under the Top-Tier AI Global HRD Invitation Program (RS-2025-25461932), supervised by the Institute for Information \& Communications Technology Planning \& Evaluation (IITP).
This work was also supported by IITP under the Artificial Intelligence Innovation Human Resources Development (IITP-RS-2026-25547954) grant funded by the Korea government (MSIT).

\bibliography{custom}

\appendix
\section{Full Perception-Mediated Results}\label{app:perception-details}

Table~\ref{tab:perception-detail-results} reports the full perception-mediated reasoning results, including VStar subtask and HRBench split scores used in the benchmark aggregates. V-DA and V-RP denote VStar's attribute-recognition and spatial-relationship-reasoning subtasks. The HRBench single and cross splits denote fine-grained single-instance and cross-instance perception.

\begin{table*}[t]
\centering
\small
\setlength{\tabcolsep}{8pt}
\begin{tabular}{lrrrr}
\toprule
Metric & Base & Direct RL & OPKD + RL & PTA + RL \\
\midrule
\multicolumn{5}{c}{\textbf{VStar results}} \\
\midrule
VStar mean@4 & 59.03 & 61.65 & 59.55 & 64.53 \\
VStar best@4 & 75.39 & 73.30 & 80.10 & 78.01 \\
V-DA mean@4 & 57.39 & 60.00 & 58.91 & 65.65 \\
V-DA best@4 & 71.30 & 69.57 & 74.78 & 73.91 \\
V-RP mean@4 & 61.51 & 64.14 & 60.53 & 62.83 \\
V-RP best@4 & 81.58 & 78.95 & 88.16 & 84.21 \\
\midrule
\multicolumn{5}{c}{\textbf{HRBench4K split results}} \\
\midrule
H4 mean@4 & 62.88 & 65.25 & 62.25 & 66.63 \\
H4 best@4 & 64.50 & 70.00 & 65.50 & 68.50 \\
H4 cross mean@4 & 50.75 & 54.50 & 52.00 & 53.00 \\
H4 cross best@4 & 52.00 & 60.00 & 54.00 & 55.00 \\
H4 single mean@4 & 75.00 & 76.00 & 72.50 & 80.25 \\
H4 single best@4 & 79.00 & 80.00 & 78.00 & 82.00 \\
\midrule
\multicolumn{5}{c}{\textbf{HRBench8K split results}} \\
\midrule
H8 mean@4 & 54.13 & 58.88 & 53.00 & 59.63 \\
H8 best@4 & 61.00 & 61.00 & 56.00 & 63.50 \\
H8 cross mean@4 & 44.00 & 48.50 & 43.75 & 51.50 \\
H8 cross best@4 & 52.00 & 50.00 & 47.00 & 55.00 \\
H8 single mean@4 & 64.25 & 69.25 & 62.25 & 67.75 \\
H8 single best@4 & 70.00 & 75.00 & 67.00 & 74.00 \\
\midrule
\multicolumn{5}{c}{\textbf{Aggregate results}} \\
\midrule
Macro mean@4 & 58.68 & 61.92 & 58.27 & 63.59 \\
Macro best@4 & 66.96 & 68.10 & 67.20 & 70.00 \\
Weighted mean@4 & 58.56 & 62.02 & 57.83 & 63.27 \\
Weighted best@4 & 64.10 & 66.33 & 62.81 & 67.28 \\
\bottomrule
\end{tabular}
\caption{Full perception-mediated reasoning results.}\label{tab:perception-detail-results}
\end{table*}

The detailed results refine the benchmark-level patterns in Section~\ref{sec:perception}. On VStar, PTA + RL exceeds OPKD + RL in mean@4 for both attribute and spatial reasoning, while OPKD + RL remains higher in best@4 for both subtasks. This explains why their aggregate ordering changes between mean@4 and best@4.

On HRBench4K, PTA + RL leads in benchmark-level mean@4 and both single-image metrics, while Direct RL leads in benchmark-level best@4 and both cross-image metrics. On HRBench8K, PTA + RL leads in both benchmark-level and cross-image metrics, including a cross-image mean@4 increase from 43.75 to 51.50 over OPKD + RL. Direct RL remains higher on the two single-image metrics. The advantage at 8K is therefore concentrated in cross-image perception at the higher resolution.

\section{Significance of the Benchmark-Level Differences}
\label{app:significance}

\begin{table}[t]
\centering
\small
\begin{tabular}{lrrrr}
\toprule
Benchmark & $n$ & $\Delta$ & $z$ & $p$ \\
\midrule
NQ & 3{,}610 & $+0.41$ & $0.38$ & $0.70$ \\
PopQA & 14{,}267 & $+1.61$ & $2.86$ & $0.004$ \\
HotpotQA & 7{,}405 & $-0.15$ & $-0.22$ & $0.83$ \\
Musique & 2{,}417 & $+2.73$ & $3.23$ & $0.001$ \\
VStar & 191 & $+4.98$ & $1.00$ & $0.32$ \\
HRBench4K & 800 & $+4.38$ & $1.83$ & $0.067$ \\
HRBench8K & 800 & $+6.63$ & $2.68$ & $0.007$ \\
\bottomrule
\end{tabular}
\caption{Benchmark-level mean@4 differences between PTA + RL and OPKD + RL, with the number of evaluation questions $n$. $\Delta$ is in percentage points.}\label{tab:significance}
\end{table}

Table~\ref{tab:significance} tests the benchmark-level mean@4 differences between PTA + RL and OPKD + RL reported in Tables~\ref{tab:retrieval-results}~and~\ref{tab:perception-results}. A question's mean@4 score is the average over its four sampled rollouts. We bound the variance of that score by $\hat{p}(1-\hat{p})$, where $\hat{p}$ is the benchmark accuracy of the system, and compute
\[
  z = \frac{\Delta}{\sqrt{\bigl(\hat{p}_1(1-\hat{p}_1) + \hat{p}_2(1-\hat{p}_2)\bigr)/n}}
\]
from the reported accuracies and the number of evaluation questions. The statistic treats questions as independent and does not pair the two systems on the same question. Reported $p$-values are two-sided and unadjusted for multiple comparisons. Each setting is trained once, so the variability captured here is that of evaluation sampling rather than of training seeds.

Three differences reach $p < 0.01$. Musique and HRBench8K are the multi-hop retrieval and higher-resolution visual settings in which Sections~\ref{sec:retrieval}~and~\ref{sec:perception} locate the largest gains. PopQA has the largest evaluation set of the seven. HRBench4K reaches $p < 0.1$. Both the variance bound and the unpaired form are conservative, so these values are obtained under a test that understates the evidence. The differences on NQ, HotpotQA, and VStar are not separable from zero at this resolution, and VStar has 191 evaluation questions, the fewest of the seven.

\FloatBarrier
\section{Teacher-Conditioned Rollout Construction}
\label{app:guidance}

Tool-use distillation can expose the teacher to behavioral information that the student does not observe at inference time. A teacher-only system instruction may, for example, specify when to call a tool, when to stop gathering observations, or which evidence to prioritize. In PTA, such information enters the rollout through two coupled channels. It shapes the teacher's next-token distribution, and it also shapes which student-proposed content is accepted or corrected.

We denote this teacher-side information by a signal $\psi$. The student does not receive $\psi$ at inference time, but the teacher verifier can be conditioned on it during training. With $\psi$ provided, the verifier returns
$c^{\mathrm{v},\psi}_{t,k} = V_\phi(\tilde{c}_{t,k} \mid \tau_t^\psi, R_{t,k-1}^\psi; \psi)$.

As the verifier's $\psi$-conditioned decisions accumulate, they determine which committed assistant turns $A_t^\psi$ are formed, which tool calls are executed, and which contexts are used for later KL supervision. The student is aligned through the committed trajectory built under the $\psi$-conditioned verifier and the teacher distribution evaluated on that trajectory, without ever observing $\psi$ at inference time.

The reported experiments hold $\psi$ fixed. The tool-call replacement ablation leaves it unchanged, and we do not measure its effect separately.

Giving the teacher information the student never observes follows privileged learning \citep{vapnik2015learning, lopezpaz2016unifying}, and distillation transfers the behavior it produces \citep{zhao2026selfdistilled}. In those methods the information shifts the teacher's predictions over a fixed trajectory. In PTA it conditions the verifier, so it also decides which trajectory is committed at all.

\section{The Committed-Token Distribution}
\label{app:committed-distribution}

This appendix states the distribution that the teacher verifier induces over committed tokens, and its relation to the student policy. It formalizes the token-level rule of Section~\ref{sec:rollout-semantics} and makes precise in what sense a PTA rollout is student-induced.

\paragraph{Setup.}
Consider one position inside a proposed chunk. Let $h$ denote the verified context at that position, consisting of the committed trajectory $\tau_t$, the pending assistant turn $R_{t,k-1}$, and the tokens already verified earlier in the same chunk. Write $p_\theta(\cdot \mid h)$ for the student conditional and $q_\phi(\cdot \mid h; \psi)$ for the teacher conditional under teacher-side guidance $\psi$. Let
\[
  \mathcal{T}_K(h) = \text{the $K$ tokens with largest } q_\phi(\cdot \mid h; \psi)
\]
be the teacher's top-$K$ support, and let $x^\star(h) = \arg\max_x q_\phi(x \mid h; \psi)$ be its top-1 token, so that $x^\star(h) \in \mathcal{T}_K(h)$ for every $K \ge 1$.

\paragraph{The verifier as a map.}
The student proposes $\tilde{x} \sim p_\theta(\cdot \mid h)$. The verifier keeps the proposal when it lies in the teacher's top-$K$ support and replaces it with the teacher's top-1 token otherwise,
\[
  V_\phi(\tilde{x} \mid h) =
  \begin{cases}
    \tilde{x} & \tilde{x} \in \mathcal{T}_K(h), \\
    x^\star(h) & \text{otherwise}.
  \end{cases}
\]
Given $h$, this is a deterministic map, so the law of the committed token is the pushforward of the student conditional under $V_\phi$.

\paragraph{Induced distribution.}
Let
\[
  r(h) = \sum_{y \notin \mathcal{T}_K(h)} p_\theta(y \mid h)
\]
be the student probability mass that the teacher rejects at $h$. Writing $\mathcal{T}$ for $\mathcal{T}_K(h)$ and $x^\star$ for $x^\star(h)$, the committed token follows
\[
  \pi(x \mid h) =
  \begin{cases}
    p_\theta(x \mid h) & x \in \mathcal{T} \setminus \{x^\star\}, \\
    p_\theta(x^\star \mid h) + r(h) & x = x^\star, \\
    0 & x \notin \mathcal{T},
  \end{cases}
\]
which is normalized because $\sum_{y \in \mathcal{T}_K(h)} p_\theta(y \mid h) + r(h) = 1$. Writing $\bar{p}_\theta(\cdot \mid h)$ for the student conditional restricted to $\mathcal{T}_K(h)$ and renormalized, the same distribution is a two-component mixture,
\[
  \pi(\cdot \mid h) = \bigl(1 - r(h)\bigr)\, \bar{p}_\theta(\cdot \mid h) + r(h)\, \delta_{x^\star(h)}.
\]
With probability $1 - r(h)$ the committed token is the student's own proposal, drawn from the student law truncated to the teacher's top-$K$ support. With probability $r(h)$ it is the teacher's top-1 token. This is the hybrid character of a PTA rollout in its most compact form.

\paragraph{Replacement rule.}
The point mass in this mixture is the teacher's top-1 token, which pairs a deterministic replacement with the deterministic acceptance test. Replacing instead by a draw from the teacher conditional would give
\[
  \pi'(\cdot \mid h) = \bigl(1 - r(h)\bigr)\, \bar{p}_\theta(\cdot \mid h) + r(h)\, q_\phi(\cdot \mid h; \psi),
\]
so the two rules differ only in this term.

\paragraph{Distance from the student policy.}
The two laws differ only at $x^\star(h)$, where $\pi$ places an additional $r(h)$, and on the rejected support, where $\pi$ places nothing. Summing these differences gives
\[
  \bigl\| \pi(\cdot \mid h) - p_\theta(\cdot \mid h) \bigr\|_{\mathrm{TV}} = r(h),
\]
so the committed-token law departs from the student policy by exactly the mass the teacher rejects at that position, and by nothing else. The realized replacement frequency, whose expectation is $\mathbb{E}[r(h)]$, is the complement of the retention rates in Table~\ref{tab:verifier-stats}: 9.5\% of committed assistant tokens in retrieval and 11.6\% in perception, and 0.5\% and 7.0\% within tool-call spans.

\paragraph{The role of $K$.}
The support size interpolates between two familiar regimes. When $\mathcal{T}_K(h)$ is the full vocabulary, no proposal can be rejected, $r(h) = 0$, and $\pi = p_\theta$, so the rollout is purely on-policy. When $K = 1$, the teacher's top-1 token is the only admissible one, $r(h) = 1 - p_\theta(x^\star(h) \mid h)$, and $\pi = \delta_{x^\star(h)}$, so the committed sequence is the teacher's greedy continuation and the rollout is off-policy. PTA uses $K = 3$ (Appendix~\ref{app:training-setup}), and the measured values of $r$ place it near the on-policy end of this range.

\paragraph{Trajectory level.}
Committed tokens are produced by applying $\pi$ at each position, conditioned on the prefix committed so far, so a committed trajectory follows the kernel $\pi$ rather than $p_\theta$. Couple the two processes by reusing the same proposal $\tilde{x}$ at every position. They emit the same token whenever the proposal is accepted, and their contexts therefore remain identical until a proposal is rejected. Under this coupling the committed trajectory equals the student's own rollout precisely when no proposal is rejected anywhere along it, an event of probability $\mathbb{E}\bigl[\prod_i \bigl(1 - r(h_i)\bigr)\bigr]$ over the positions $i$ of the trajectory. This quantity decreases with trajectory length, which is the sense in which long-horizon tool use gives the verifier more occasions to act.

\paragraph{Supervision target.}
The mixture determines which positions the student is supervised at, not what it is supervised toward. The objective of Section~\ref{sec:objective} matches $p_\theta$ against the teacher conditional $q_\phi$ at committed assistant positions, and $q_\phi$ does not depend on $\pi$. Guidance $\psi$ enters the committed distribution only through $\mathcal{T}_K(h)$ and $x^\star(h)$, so teacher-side information reshapes the acceptance set and the replacement token without changing the form of either the objective or the mixture.

\section{Verifier Acceptance and Correction Statistics}
\label{app:verifier-stats}

\FloatBarrier
\begin{table}[ht]
\centering
\small
\begin{tabular}{@{}p{0.60\linewidth}rr@{}}
\toprule
Statistic & Retr. & Perc. \\
\midrule
Committed assistant tokens retaining the student proposal & 90.5 & 88.4 \\
Tokens within tool-call spans retaining the student proposal & 99.5 & 93.0 \\
Tool calls containing at least one replacement & 8.9 & 84.7 \\
\bottomrule
\end{tabular}
\caption{Verifier acceptance and correction rates, in percent, aggregated over the training logs of the alignment runs used in the paper.}\label{tab:verifier-stats}
\end{table}

\begin{figure*}[t]
\centering
\includegraphics[width=\textwidth]{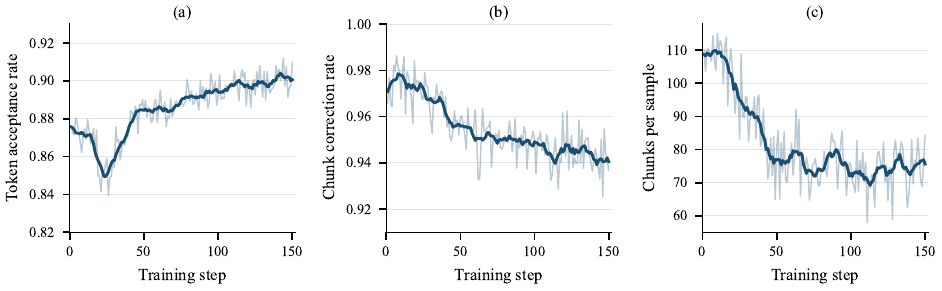}
\caption{Verifier behavior over the first 150 training steps. (a) Fraction of student-proposed tokens that the verifier keeps. (b) Fraction of chunks in which at least one token is replaced. (c) Number of chunks accumulated per sample, each of which corresponds to one call to the teacher verifier in Algorithm~\ref{alg:pta-rollout}. Light lines are per-step values and solid lines are a nine-step moving average.}\label{fig:verifier-rates}
\end{figure*}

Table~\ref{tab:verifier-stats} aggregates the training logs of the alignment runs used in the paper. Token-level and call-level rates describe different quantities and should be read separately. In the perception environment, 93.0\% of tool-call tokens keep the student proposal while 84.7\% of tool calls contain at least one replacement, so verifier intervention is distributed across many calls rather than concentrated in a few. Three factors combine to produce this pattern: perception tool-call spans are longer, at 58.3 against 29.4 tokens on average, a call counts as corrected once a single token is replaced, and the in-span correction rate is higher than in retrieval. Among corrected calls, the median number of replaced tokens is five in perception and one in retrieval.

The cross-environment difference follows the structure of the tool arguments. Natural-language search queries admit many semantically equivalent expressions, whereas bounding-box arguments are numerical coordinates that identify a specific region and therefore admit less variation. Replacements inside tool-call spans account for 0.05\% of all committed assistant tokens in retrieval and 0.82\% in perception.

Figure~\ref{fig:verifier-rates} resolves the perception rates over training. Averaged over the first ten steps, the verifier keeps 0.873 of the student's proposed tokens, and over the last ten steps it keeps 0.902. The complement of this quantity is the rejected mass $r(h)$ of Appendix~\ref{app:committed-distribution}, which is exactly the total variation distance between the committed-token distribution and the student policy. It falls from 0.127 to 0.098, a reduction of 23\%, so the committed rollout moves closer to the student's own policy as distillation proceeds.

The chunk-level rate moves in the same direction but stays much higher. The fraction of chunks receiving at least one replacement declines from 0.975 to 0.941, so nearly every chunk is still corrected somewhere while roughly one token in ten is replaced. Correction is spread thinly across the rollout rather than concentrated in a few chunks.

Chunks per sample fall from 109 to 77 over the same window. A chunk ends at its first rejected token, so a higher acceptance rate yields longer accepted runs and fewer chunks. Because each chunk is one call to the teacher verifier, the teacher-side cost of constructing a rollout decreases as training proceeds.

These statistics report the frequency and location of intervention. They do not classify how each individual correction changes the semantics of a tool call.

\section{Teacher-Gated Carryover under Student Updates}
\label{app:teacher-gated-carryover}

Lookahead scheduling can carry an unfinished rollout across trainer steps. This creates a possible stale-prefix issue: a carryover state may have been produced under the pre-update student parameters $\theta_s$, but continued after the update by $\theta_{s+1}$. PTA does not claim trajectory equivalence between such a carryover rollout and a rollout regenerated from $\theta_{s+1}$. Instead, it guarantees that raw student drift after the update cannot directly enter the rollout.

Let a carryover state be
\[
S_s=(\tau_s,R_s),
\]
where $\tau_s$ is the committed trajectory and $R_s$ is the pending assistant turn. A carryover state is PTA-consistent if it satisfies three conditions. First, every assistant span in $\tau_s$ is a finalized committed assistant turn. Second, every observation in $\tau_s$ was inserted only after a committed assistant turn was parsed as a valid tool call and the environment transition completed. Third, $R_s$ is either empty or consists only of teacher-verified chunks. In particular, $S_s$ contains no raw speculative chunk. Carryover also never occurs between a committed tool call and its observation, since the two form a single atomic transition, so no carryover state holds a committed tool call whose observation has not yet been inserted.

After the parameter update $\theta_s \rightarrow \theta_{s+1}$, the carryover state $S_s$ is not modified. The updated student may continue generation from this state:
\[
\tilde{c}_{s+1,k} \sim p_{\theta_{s+1}}(\cdot \mid \tau_s, R_s).
\]
However, the speculative chunk $\tilde{c}_{s+1,k}$ is not appended to the pending assistant turn, the committed trajectory, the environment, or the loss target. It must first pass the same fixed teacher verifier:
\[
c^{\mathrm v}_{s+1,k}
=
V_\phi(\tilde{c}_{s+1,k}\mid \tau_s,R_s).
\]
Only the verified chunk is retained:
\[
R_{s+1,k}=R_s \oplus c^{\mathrm v}_{s+1,k}.
\]

Thus, even if the updated student produces a continuation that drifts from the teacher, that raw continuation cannot directly affect the rollout. The fixed teacher verifier applies the same accept-or-correct gate before and after the student update. If the pending assistant turn is later finalized, the resulting committed assistant turn is still composed only of teacher-verified chunks. If that committed assistant turn is parsed as a valid tool call, the returned observation is appended only after the environment transition completes.

\paragraph{Proposition.}
Lookahead carryover preserves PTA commitment semantics across student updates. A carryover rollout may be stale with respect to the updated student, but every post-update continuation must pass the fixed teacher verifier before it can affect the pending assistant turn, committed trajectory, environment transition, or distillation target.

\paragraph{Proof.}
The carryover state $S_s=(\tau_s,R_s)$ is PTA-consistent by construction, because lookahead permits carryover only after teacher verification or after a completed environment transition. The parameter update changes $\theta_s$ to $\theta_{s+1}$, but does not insert new content into $S_s$. Any new content generated after the update is first a speculative chunk. PTA retains this chunk only after applying $V_\phi$, and appends only the verified chunk to $R_s$. Therefore, the pending assistant turn remains teacher-verified. Turn finalization, trajectory commitment, and observation insertion then follow the same PTA rules as in the main rollout procedure. By induction over post-update generation steps, all states reached from a carryover state remain PTA-consistent.

\section{Training Setup}
\label{app:training-setup}

All experiments were run on a single node with four GPUs. Training used a PPO-based learning loop with GRPO-style advantage estimation~\citep{schulman2017ppo,shao2024deepseekmath}, together with an asynchronous distillation setup that combines student-model optimization with teacher-model inference. The teacher model was served on the same node with four inference replicas. We did not run a separate validation stage during training; the submission configuration keeps only the training path.

Optimization used a learning rate of $1\times10^{-6}$ and a policy-update mini-batch size of 32. We did not use KL reward, KL loss, or entropy regularization during training; the entropy regularization coefficient was set to 0. Checkpoints were saved periodically.

Distillation used a forward-KL objective over a truncated teacher distribution that retains only the highest-probability candidates. All experiments used the top 256 candidates, and the speculative distillation chunk size was 128. During the teacher-student alignment stage, verification used the top 3 candidates.

Rollouts were generated asynchronously and allowed multi-turn interaction. Both student rollout generation and teacher inference were run under fixed token budgets chosen for memory efficiency.

\subsection{Search-R1-like QA Setup}

The Search-R1-like experiments used Qwen3-1.7B as the student model and Qwen3-32B as the teacher model~\citep{qwen2025qwen3}. Training used a preprocessed Search-R1-style question-answering dataset. The maximum prompt length was 8,192 tokens, the maximum response length was 16,384 tokens, and the training batch size was 32.

This setting assumes multi-turn question answering with a search tool. Each trajectory generated one sample. Each episode allowed up to 15 assistant turns and 15 user turns. At most one tool call was allowed at a time, and each tool response was limited to 2,048 tokens. Speculative distillation allowed up to 256 chunks per sample.

Rewards were configured to use one of two scoring modes. The default mode used exact-match scoring, while an optional mode used LLM-as-a-judge scoring. In the submission training configuration, rewards were computed only on the training data without validation. The total number of training epochs was 5.

\subsection{DeepEyes Setup}

The DeepEyes experiments used Qwen3-VL-2B-Thinking as the student model and Qwen3-VL-32B-Thinking as the teacher model~\citep{bai2025qwen3vl}. Training used the DeepEyes visual-toolbox dataset. The maximum prompt length and maximum response length were both 8,192 tokens, and the training batch size was 32.

This setting assumes image-based multi-turn tool use, where the model uses a zoom-in tool to inspect local visual information. Each episode allowed up to 15 assistant turns and 15 user turns, with at most one concurrent tool call. Speculative distillation allowed up to 128 chunks per sample.

The reward combined answer accuracy, format consistency, and tool-use behavior. Answer correctness was judged with a Gemini-based judge. The total number of training epochs was 1.

\end{document}